\documentclass{article}
\usepackage{spconf,amsmath,graphicx}
\usepackage{xfrac}

\def\x{X}
\def\a{\beta}
\newcommand{\eq}[1]{Eq.~(\ref{#1})}
\newcommand{\fig}[1]{Fig.~\ref{#1}}

\usepackage{graphicx}
\usepackage{amsmath}
\usepackage{bbm}

\title{AN ALTERNATIVE MATTING LAPLACIAN}
\name{Fran\c{c}ois Piti\'e\thanks{This work is supported by the SFI research
    centre ADAPT. {\tt http://adaptcentre.ie/}}}

\address{School of Engineering \\
  Trinity College Dublin}

\begin{document}
\maketitle
\begin{abstract}
  Cutting out and object and estimate its transparency mask is a key task in
  many applications. We take on the work on closed-form matting by Levin et
  al.\cite{Levin08a}, that is used at the core of many matting techniques, and
  propose an alternative formulation that offers more flexible controls over the
  matting priors. We also show that this new approach is efficient at upscaling
  transparency maps from coarse estimates.
\end{abstract}

\section{Introduction}
\label{intro}

Pulling a matte from a film or video sequence is the act of cutting out an
object from its background by creating a transparency mask or $\alpha$ matte,
that ranges between 0 and 1. An $\alpha$ value of 0 at pixel $i$ means that only
the background colour $B_i$ is visible, an $\alpha$ value of 1 means that only
the foreground colour $F_i$ is visible, and values in-between 0 and 1 manifest a
blend between the object and the background:
\begin{equation}\label{eq:matting}
  C_i = \alpha_i F_i + (1 - \alpha_i) B_i
\end{equation}

To help the algorithms, techniques typically ask the user to scribble a {\em
  trimap}, defining foreground, background and the unknown region to be pulled.

Since the successful work of Chuang et al.~\cite{Chuang01} on bayesian matting
in 2001, the notion of matting as an inference problem has been explored by
several authors (see in particular Poisson Matting [2] and Inference Matting
[3]). In 2007 Levin et al.~\cite{Levin08a} proposed a number of remarkable
advances by finding a closed-form solution to the matting problem. Their
solution is fast and very often used as a core component in subsequent matting
techniques~\cite{He11,Shahrian13,Karacan2015ICCV}.

A key contribution from Levin et al.~\cite{Levin08a} is to consider that the
transparency value can be approximated as a linear combination of the colour
components:
\begin{equation}\label{eq:linearmattingequation}
  \alpha_i = a_i^T C_i + b_i
\end{equation}
with $a_i = [a_i^R, a_i^G, a_i^B]$. Instead of a non-linear problem with 6
unknowns (3 for $F$, 3 for $B$), we are left with a linear model with 4 unknowns
(3 for $a$ and 1 for $b$). This linear approximation is justified by the fact
that colours often spread along 3D lines in the colour space (see Hillman et
al.~\cite{hillman05}).

Interestingly this approach has been used routinely in video post-production for
a long time. For instance, tools used to extract transparency maps for green or
blue screens (also called keyers) are often based on a simple linear fit.

The second contribution of Levin's work is to introduce a smoothness constraint
that yields a global closed-form solution. They state that the matting model
should hold for overlapping $3\times3$ image patch $w_i$. This leads to the
following energy to minimise:
\begin{equation}\label{eq:cfm:energy}
  J(\alpha, a, b) = \sum_i \sum_{j\in w_i } \left(a_i^T C_j + b_i -
  \alpha_j\right)^2 + \epsilon \|a_j\|^2
\end{equation}
The term $\epsilon \|a_j\|^2$ adds further smoothness to the matte by damping
$a$ towards zero and thus flattening the matte.

At this point Levin et al have chosen to integrate $a$, $b$ out of the problem
and so generate a marginalised estimate of $\alpha$. This leads to a linear
system as follows:
\begin{equation}
  (L + D) \alpha = D \alpha_0
\end{equation}
where $\alpha_0$ is the initial guess for the transparency values, and $D$ a
diagonal matrix of the confidence values for this guess. The sparse matrix $L$
is the matting Laplacian, derived from \eq{eq:cfm:energy}, with entries as
follows:
\begin{equation}\label{eq:cfm:weights}\small
  L_{i,j} = \sum_{k|i,j\in w_k} \delta_{i,j} - \frac{1 + (C_i-\mu_k) \left( R_k
    + \frac{\epsilon}{|w_k|}\right)^{-1}(C_j-\mu_k)}{|w_k|}
\end{equation}
where $\mu_k$ and $R_k$ are the mean and covariance of patch $w_k$.

\paragraph*{Contribution.}

\begin{figure}
  \setlength{\tabcolsep}{.5pt} \renewcommand{\arraystretch}{.7}
  \begin{tabular}{ccc}
    \includegraphics[width=0.33\linewidth]{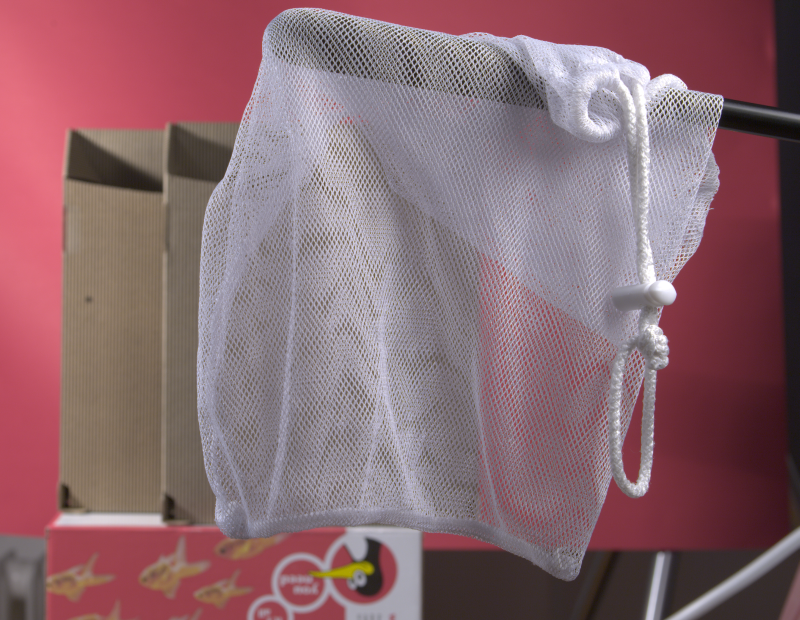} &
    \includegraphics[width=0.33\linewidth]{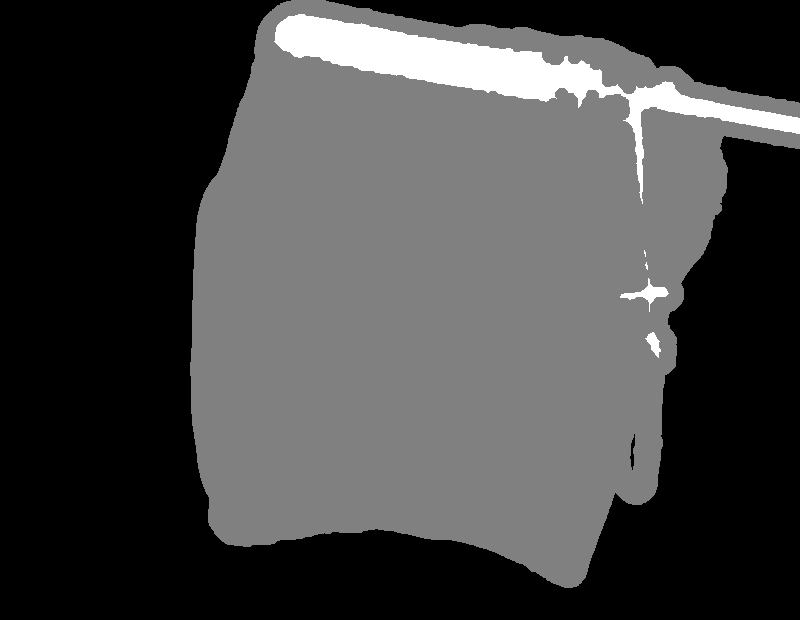} &
    \includegraphics[width=0.33\linewidth]{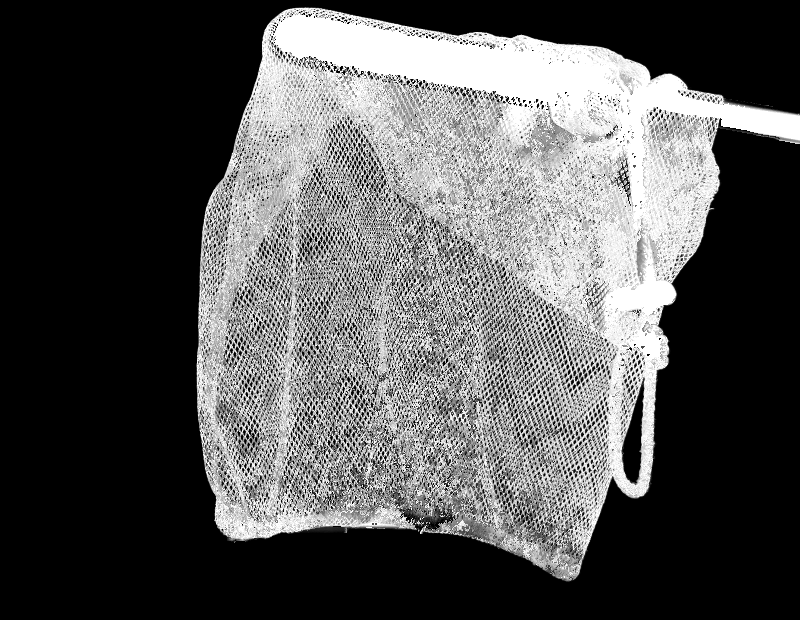}
    \\ (a) & (b) & (c)
    \\ \includegraphics[width=0.33\linewidth]{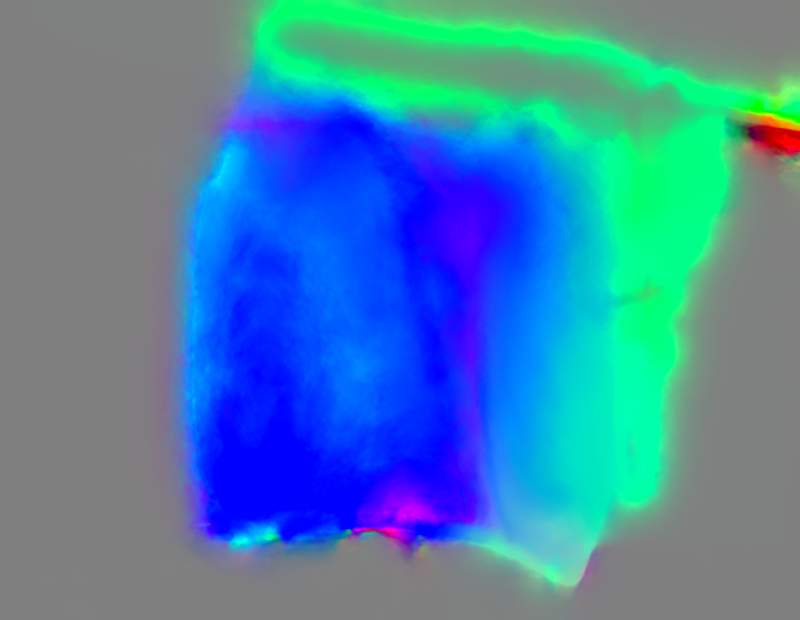}
    &
    \includegraphics[width=0.33\linewidth]{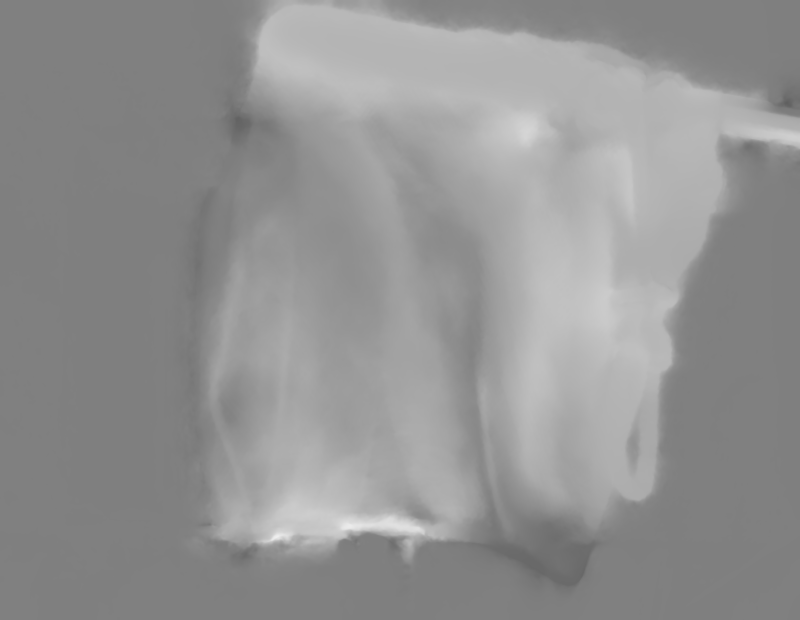}
    &
    \includegraphics[width=0.33\linewidth]{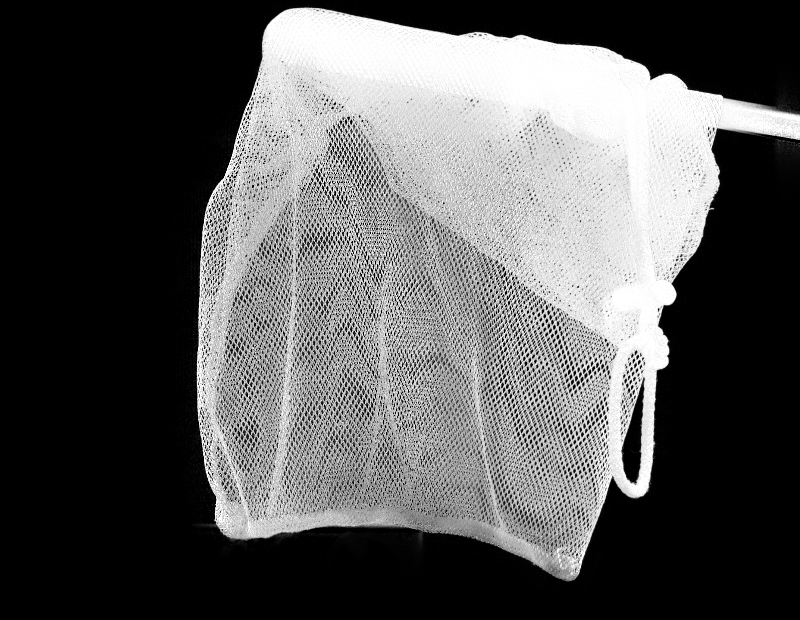}
    \\ (d) & (e) & (f) \\
  \end{tabular}
  \caption{Example. (a) input image, (b) trimap, (c) $\alpha_0$
    from~\cite{He11}, (d) estimated $a$, (e) estimated $b$, (f)
    estimated $\alpha$. }
  \label{fig:example}
\end{figure}

We argue in this paper that instead of integrating out $a$ and $b$, it might be
more interesting to integrate out $\alpha$ and solve for $a$ and $b$ instead. In
\fig{fig:example} is an example of our estimates for $a$ and $b$ ($a$ has been
rescaled to $a/5+.5$ to better show the effect). We can think of $a$ as the
colour of the colour filter applied to the input picture to reveal
$\alpha$. Whereas the values of $\alpha$ might vary quickly and change across
the image scales, the values of $a$ and $b$ are locally smooth. Levin et
al. take advantage of this for upscaling $\alpha$ map without solving the linear
system. Estimates of $a$ and $b$ are derived from the lowres $\alpha$ map and a
highres $\alpha$ map is obtained by applying \eq{eq:linearmattingequation} on
the upscaled $a$ and $b$. What we propose is to work directly with $a$ and $b$
and avoid the estimation step from $\alpha$. This will result in a cleaner
estimation of $a$ and $b$.

In closed-form matting, it is hard to finely control the scale of the spatial
smoothness. The minimum stencil size for $\alpha$ is a $5\times 5$ stencil, and
larger patches require a different approach~\cite{He10} to be practical. We show
that working directly with $a$ and $b$ allows us to use a more compact 5-point
stencil and gives us a simpler way to control the spatial smoothness.

Lastly, the matting Laplacian of \eq{eq:cfm:weights} needs to be positive
definite. In~\cite{Levin08a}, this is enforced by adding the unary prior
$\epsilon\|a\|^2$. The problem is that this prior is purely used for numerical
stability and does not yield interesting matting properties. This is noted in
Piti\'e and Kokaram~\cite{PitieK10} who argue that a better approach is to set
$\epsilon$ only in the direction of small eigenvalues of the covariance
matrix. Other priors on the patch covariance matrix are discussed
in~\cite{Singaraju09}, but then again, the analysis is limited to the patch
size. We argue in this paper that working with $a$ and $b$ yield in simpler and
more useful priors.

\paragraph*{Organisation of the paper.}
In section~\ref{sec:ourlap}, we derive an alternate matting Laplacian
formulation for $a$ and $b$. We also expose how the spatial and unary priors can
be set in this new framework. Results and comparisons of our approach to
closed-form matting are presented in section~\ref{sec:res}.

\section{Matting Colour Filter}\label{sec:ourlap}

Let us use the notation $\x = [C; 1]$ as it allows us to write the linear
matting model in a slightly more compact form:
\begin{equation}\label{eq:ourmatting}
  \alpha_i = \x_i^T \beta_i
\end{equation}
The model parameters $\beta_i = [a_i; b_i]$ is of dimension $4 \times 1$ and is
the concatenation of $a$ and $b$ of closed-form matting.

\subsection{Matting Laplacian for $\beta$}

We start from the same energy as in \eq{eq:cfm:energy}, but ignoring for the
moment the penalty term on $\epsilon \|a\|^2$ and only focusing on the spatial
constraints:
\begin{equation}
  J^*(\alpha,\a) =  \sum_i \sum_{j\in w_i } \left( \x_j^T \beta_i -
  \alpha_j\right)^2
\end{equation}
Denote $w^{-1}_j = \{i : j \in w_i\}$ and reorder the summation:
\begin{equation}
  J^*(\alpha,\a) = \sum_j \sum_{i \in w^{-1}_j} \left( \x_j^T \beta_i -
  \alpha_j\right)^2
\end{equation}
Now let us integrate out $\alpha$:
\begin{equation}
  J^*(\a) =  \sum_j \min_{\alpha_j} \sum_{i \in w^{-1}_j} \left( \x_j^T \beta_i -
  \alpha_j\right)^2
\end{equation}
The least square estimate for $\alpha_j$ is $\frac{1}{|w^{-1}_j|}\sum_{k \in w^{-1}_j} \x_j^T \beta_k$,
\begin{equation}
  J^*(\a) =  \sum_j \sum_{i \in w^{-1}_j } \left( \x_j^T \beta_i - \frac{1}{|w^{-1}_j|}\sum_{k \in w^{-1}_j} \x_j^T \beta_k \right)^2
\end{equation}
\begin{equation}
  J^*(\a) = \sum_j \sum_{i \in w^{-1}_j } \left(
  \frac{1}{|w^{-1}_j|}\sum_{k \in w^{-1}_j} \x_j^T (\beta_i - \beta_k)
  \right)^2
\end{equation}
This is a quadratic form in $\a$. We can rewrite it as follows:
\begin{equation}
  J^*(\a) = \sum_{i,k} \beta_i^T A_{i,k} \beta_k
\end{equation}
with
\begin{equation}
  A_{i,k}  =
  \begin{cases}
    - \sum_{j : i,k \in w^{-1}_j} \frac{2}{|w^{-1}_j|^2} \x_j \x_j^T &
    \text{if } i \neq k \\ - \sum_{l \neq i} A_{i,l} & \text{if } i = k
  \end{cases}
\end{equation}
Similarly to the closed-form approach, we end up with a linear system
of equations:
\begin{equation}
  \left(A + A_0\right)\a = A_0 \beta_0
\end{equation}
The trimap and other priors are encoded into $A_0$ and $\beta_0$ and will be
discussed in the following sections on unary priors.

This time the matting Laplacian $A=(A_{i,k})$ is a block banded matrix, with
each block matrix being of size $4 \times 4$.  Importantly this linear system
solves the exact same energy as in closed-form matting. Thus solving for $\beta$
and then resolving the matting equation $\alpha=x_i^T\beta_i$ is the same as
solving directly for closed-form matting. In the following, we discuss the
benefits of this approach.

\subsection{Matting as an Anisotropic Diffusion}

In closed-form matting, the use of the inverse of the covariance matrix in
\eq{eq:cfm:weights} requires a minimum patch size to ensure numerical stability
of the matrix inversion. There is no such numerical issue here, and we can
safely consider much smaller patch sizes. For instance, for patches of size 2
({\em i.e.} vertical and horizontal pairs), it follows a 5-point stencil linear
system:
\begin{equation}\label{eq:mla:weights}
  \boxed{
    A_{i,k}  =
    \begin{cases}
      -  \frac{1}{2} (\x_i \x_i^T + \x_k \x_k^T)
      & \text{if }   i \neq k, k \in \mathcal{N}_i \\
      - \sum_{l \neq i} A_{i,l} & \text{if }   i = k
  \end{cases}}
\end{equation}
where $\mathcal{N}_i$ is the 5-point stencil of pixel $i$.

Interestingly this system corresponds to an anisotropic diffusion process for
$\a$. In continuous settings, the smoothness constraint is in fact equivalent
to:
\begin{equation}
  \mathrm{div}(\x^T  \nabla \a )= 0
\end{equation}
where the spatial gradient $\nabla \a$ is of dimension $4 \times 2$. Note that
it is possible to start from this formulation to derive \eq{eq:mla:weights}
using finite difference approximation.

\subsection{Spatial Priors}

In closed-form matting, the prior on the spatial smoothness is controlled via a
penalty energy $\epsilon \|a\|^2$ which biases $a$ towards zero, hence making
sure that $\alpha$ is smooth. This prior is used to enable the stable inversion
of the colour covariance matrix.

With our proposed formulation, there is another way of imposing spatial prior by
setting a prior on $\x\x^T$. The matrix $\x\x^T$ is of rank 1. This can make $A$
indefinite and thus unstable to solve. However, a number of simple priors on the
colour distribution of $\x$ can help making $A$ definite positive.

For instance, we can replace $\x_i\x_i^T$ with $\x_i\x_i^T + \epsilon_s I_4$ in
\eq{eq:mla:weights}. This increases the rank to 4 by adding a small isotropic
Gaussian prior on the distribution of $\x$. Alternatively, we can replace
$\x_i\x_i^T$ with a local estimate of $\mathbbm{E}\{\x_i\x_i^T\}$, which can be
obtained by a spatial Gaussian blur (variance $\sigma_s$) of the values of
$\x_i\x_i^T$. 

We believe that these kind of spatial priors are more principled and useful.

\subsection{Unary Priors}

In closed-form matting, the unary priors are limited to a prior on $\alpha$ and
the penalty energy $\epsilon \|a\|^2$. Here we explore what can be gained from
using a full Multivariate Gaussian prior on $\beta$:
\begin{equation}
  U_{\a}(\beta_i) = \frac{1}{2} \left(\beta_i -
  \beta_{i,0}\right)^TA_{i,0}\left(\beta_i - \beta_{i,0}\right)^T
\end{equation}
where $A_{i,0}$ is the covariance of the of MVG prior and $\beta_{i0}$ the
expected mean value. For a practical design of $A_{i,0}$, consider that we have
$N$ samples $\x_1,\cdots,\x_N$ for which we have expected
$\alpha_1,\cdots,\alpha_N$ and confidence measure $\lambda_1, \cdots,
\lambda_N$:
\begin{equation}
  \begin{aligned}
    & U_{\a}(\beta_i)
    =  \frac{1}{2N} \sum_{n=1}^N \lambda_n \left(\x_n^T \beta_i - \alpha_n \right)^2 \\
    &=  \frac{1}{2N} \sum_{n=1}^N \lambda_n  \beta_i^T \x_n \x_n^T
    \beta_i + \alpha_n^2 - 2 \lambda_n  \alpha_n \x_n^T \beta_i \\
    &= \frac{1}{2} \left(\beta_i - A_{i,0}^{-1} \mu_{i,0}\right)^T A_{i,0} \left(\beta_i - A_{i,0}^{-1} \mu_{i,0}\right)
  \end{aligned}
\end{equation}
with $A_{i,0} = \frac{1}{N}\sum_{n=1}^{N} \lambda_n  \x_n \x_n^T$ and
$A_{i,0}\beta_{i,0}=\mu_{i,0}=\frac{1}{N}\sum_{n=1}^{N} \lambda_n
\alpha_n\x_n$. The complete linear system is as follows:
\begin{equation}
  \left(A + A_{0}\right) \beta = \mu_0
\end{equation}
where $A_{0}$ is a block diagonal made up of all the $A_{i,0}$ matrices and
vector $\mu_{0} = A_{i,0} \beta_0$ is the concatenation of all $\mu_{0,i}$.

\paragraph*{Prior on $\alpha$.} If we want to for $\alpha_i$ to be closed
to $\alpha_{i0}$, then we can set $A_{i,0} = \lambda \x_i \x_i^T$ and
$\mu_{\alpha\x} = \lambda \alpha_{i0}\x_i$.

\paragraph*{Prior on $F$.} For a prior on
$\x_F$, then we set $A_{i,0} = \lambda \x_F \x_F^T$ and $\mu_{i,0} = \lambda
\x_F$. Alternatively, if we have a distribution for $\x_F$, then $A_{i,0}=
\mathbbm{E}\left\{\lambda\x_F\x_F^T\right\}$ and $\mu_{i,0} =
\mathbbm{E}\{\lambda\x_F\}$.

\paragraph*{Prior on $B$.} Similarly, if we wish to impose a prior on
$\x_B$, then we set $A_{i,0} = \lambda \x_B \x_B^T$ and $\mu_{i,0} = \lambda
\x_B$. Alternatively, we can set $A_{i,0}=
\mathbbm{E}\left\{\lambda\x_F\x_F^T\right\}$ and $\mu_{i,0} =
\mathbbm{E}\{\lambda\x_F\}$.

\begin{figure}
  \setlength{\tabcolsep}{.5pt} \renewcommand{\arraystretch}{.7}

  \begin{tabular}{ccc}
    \includegraphics[width=0.33\linewidth]{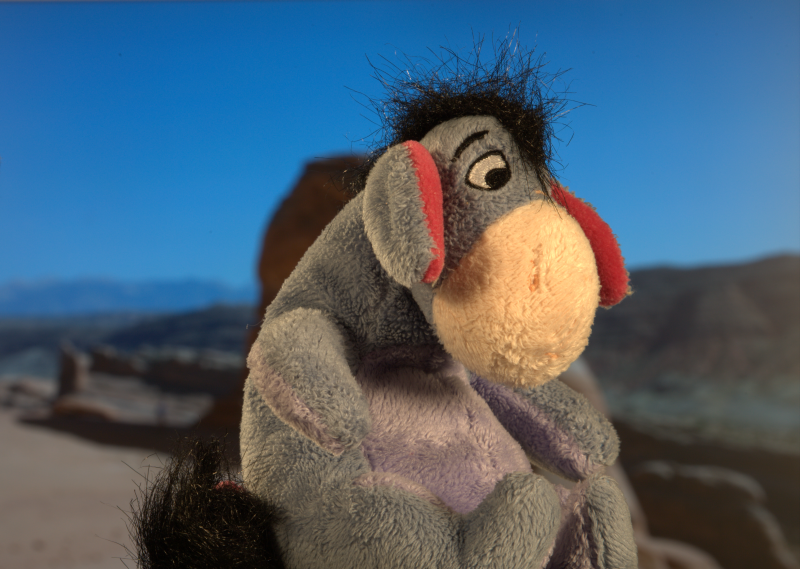} &
    \includegraphics[width=0.33\linewidth]{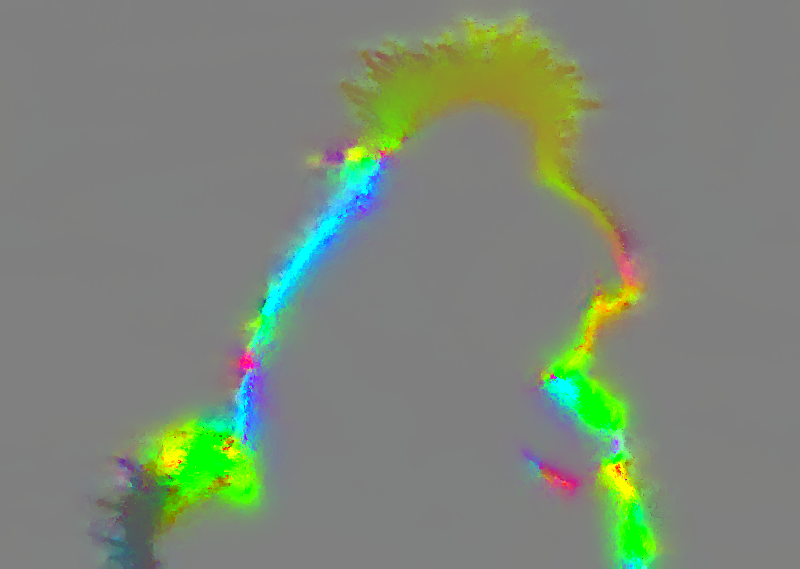}
    &
    \includegraphics[width=0.33\linewidth]{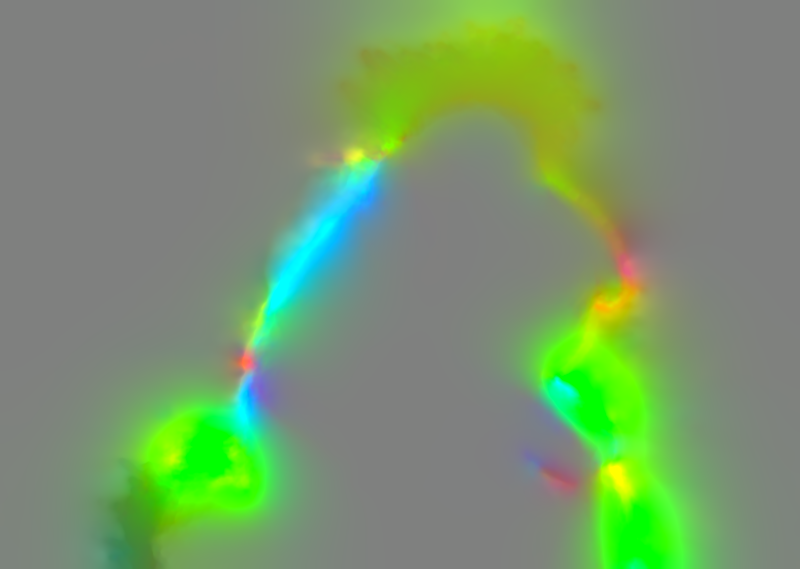}
    \\ (a) & (b) & (c)
    \\ \includegraphics[width=0.33\linewidth]{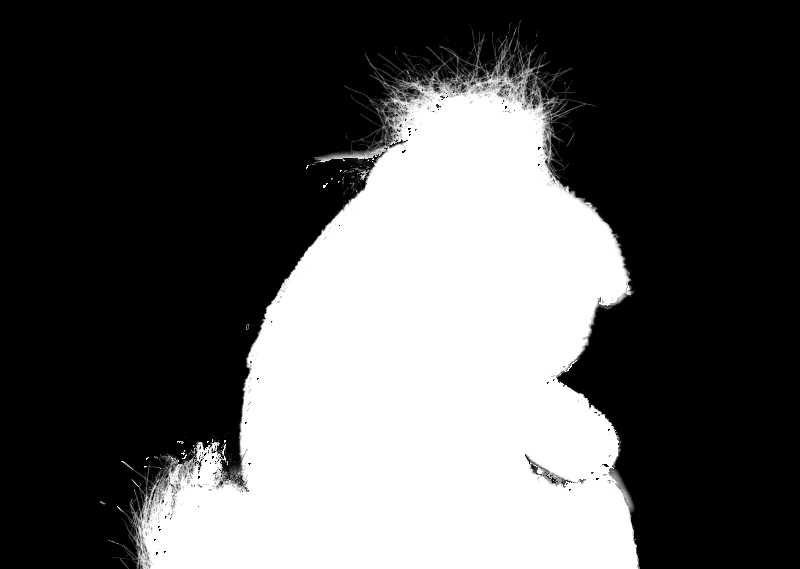}
    &
    \includegraphics[width=0.33\linewidth]{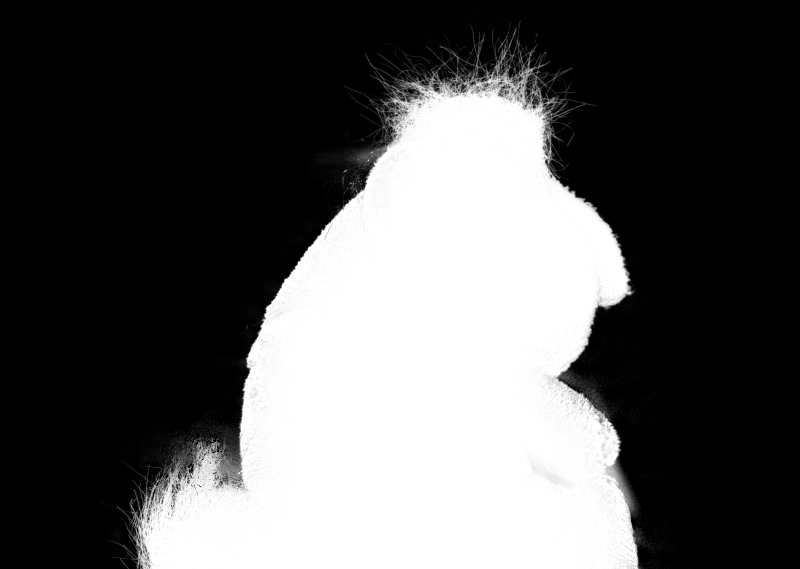}
    &
    \includegraphics[width=0.33\linewidth]{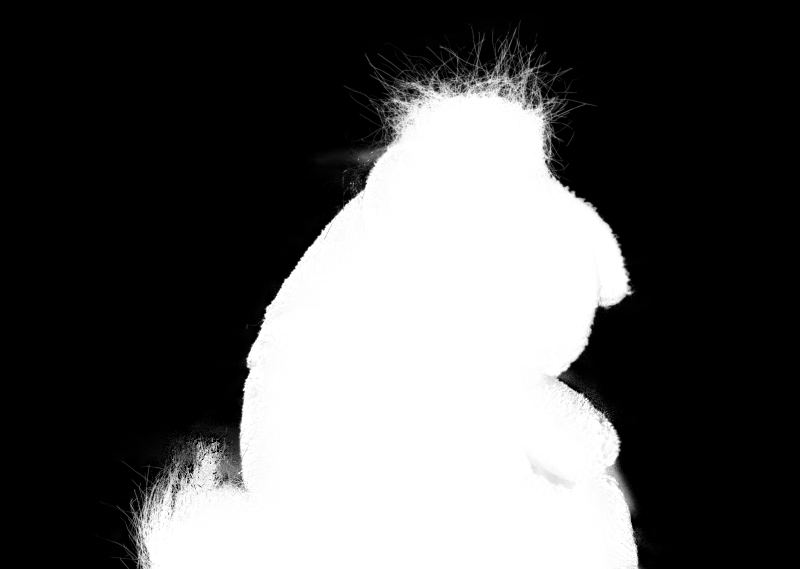}
    \\ (d) & (e) & (f) \\
  \end{tabular}
  \caption{Input image on (a). (b) $\beta$ estimated using closed-form, (c)
    using our technique ($\sigma_s=1$, $\epsilon_s=10^{-4}$), MAD(e,f)=0.0025.}
\label{fig:donkeyvsLevin}
\end{figure}

\begin{figure}
  \setlength{\tabcolsep}{.5pt} \renewcommand{\arraystretch}{.7}
  \begin{tabular}{ccc}
    \includegraphics[width=0.33\linewidth]{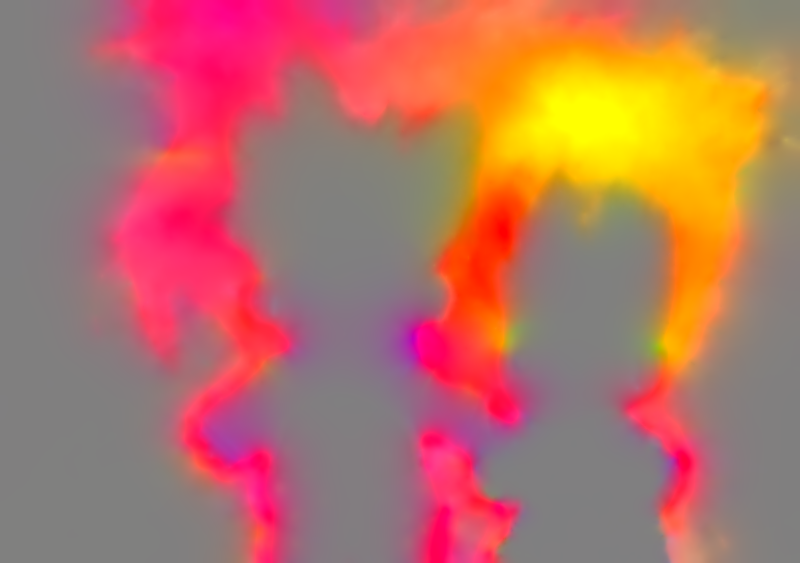}
    &
    \includegraphics[width=0.33\linewidth]{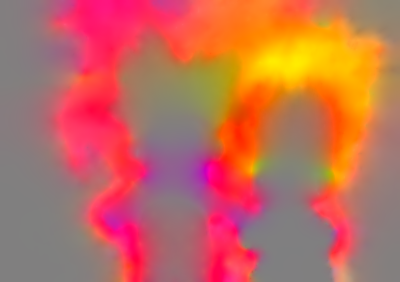}
    &
    \includegraphics[width=0.33\linewidth]{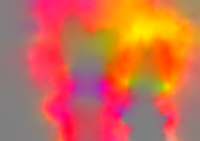}
    \\ (a) & (b) & (c)
    \\ \includegraphics[width=0.33\linewidth]{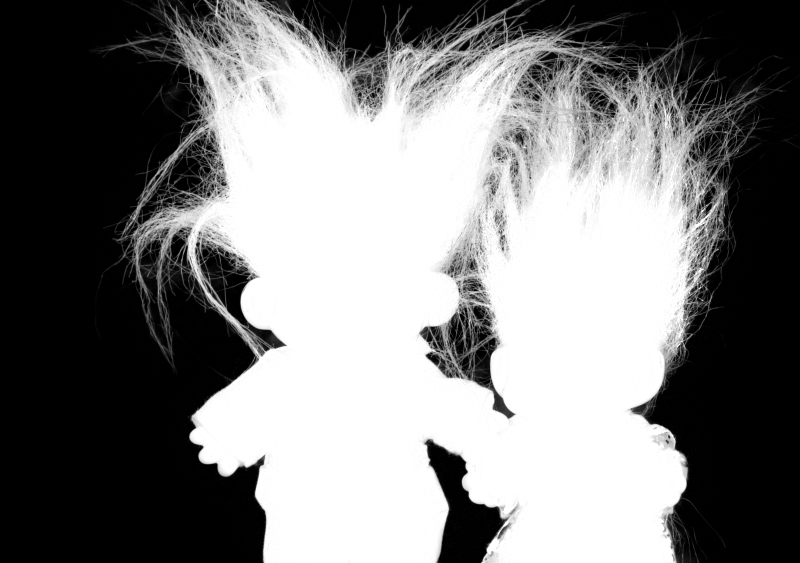}
    &
    \includegraphics[width=0.33\linewidth]{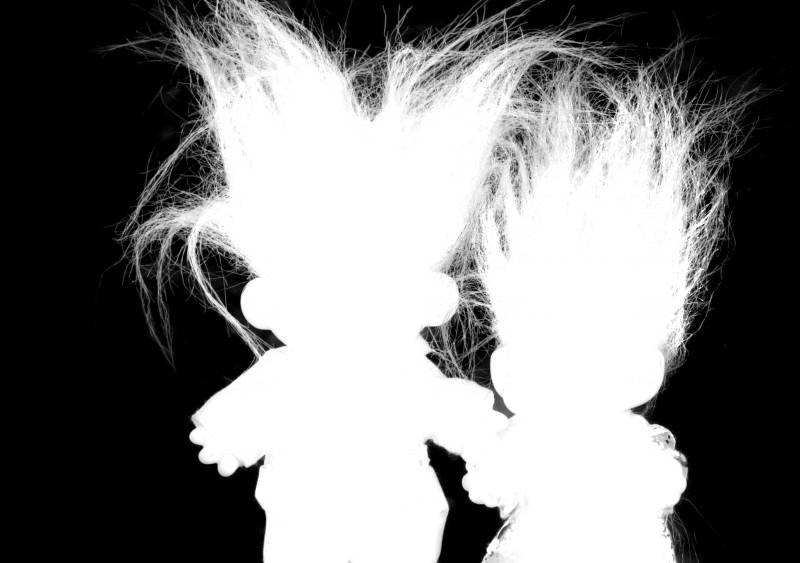}
    &
    \includegraphics[width=0.33\linewidth]{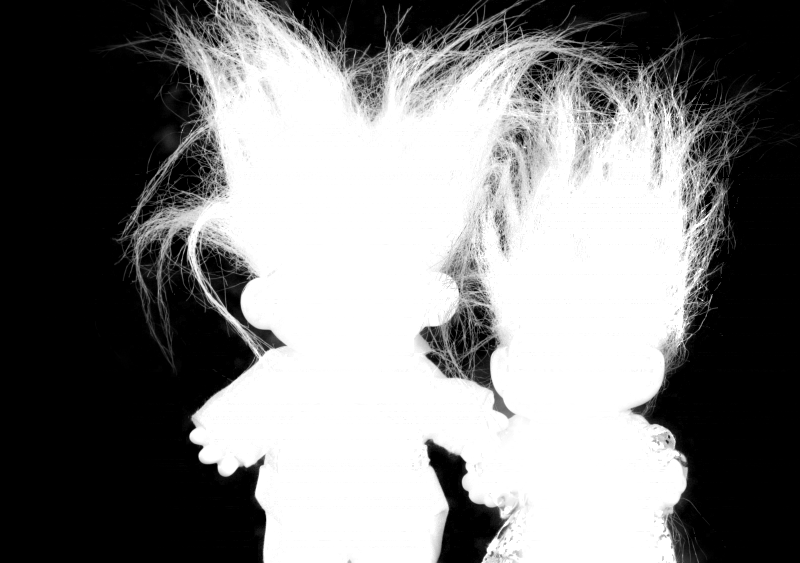}
    \\ (d) & (e) & (f) \\
  \end{tabular}
  \caption{Estimation at different scales. (a) $\beta$ estimated from full
    resolution image, $\alpha_0$ and confidence $\lambda$, (b) from half res,
    (c) from quarter res. (d) derived $\alpha$ at full res, (e) derived using
    upsampled $\beta$ from (f) (SSIM(d,e)=0.9768; MAD(d,e)=0.0085), derived
    using upsampled $\beta$ from (c) (SSIM(d,f)= 0.9544; MAD(d,f)=0.0128). }
  \label{fig:scales}
\end{figure}

\section{Results}\label{sec:res}

We expect our technique to produce transparency maps that are very similar to
the ones produced by closed-form matting. The only numerical difference with
closed-form results from the choice of spatial and unary priors.

Although closed-form matting is being extensively used as a smoothing constraint
or final filtering stage (eg. ~\cite{He11,Chen13}), it ranks poorly against the
state of the art for very sparse trimaps. We propose in the following to look at
a typical matting scenario where matting is first bootstrapped with a sampling
step~\cite{He11}, which gives us an initial $\alpha_0$ map and confidence map
$\lambda$.

\subsection{Comparison with Closed-Form Matting}
In \fig{fig:scales}, we show the difference between the $\beta$ values estimated
using closed-form matting (b) and our technique (c), with $\sigma_s=1$,
$\epsilon_s=10^{-4}$. The Mean Absolute Difference (MAD) between the derived
$\alpha$ maps (e) and (f) is 0.0025, which confirms that both approaches produce
very similar results. Comparisons for the 26 examples of the training dataset of
the alpha matting evaluation website~\cite{alphamatting.com, TUW-180666} are
presented in Table~\ref{table:diff}.

\subsection{Upscaling}
One key advantage of our approach is that the estimation of $\beta$ at a coarse
scale can be used for estimating $\alpha$ at higher scales. In \fig{fig:scales}
we show on (a) $\beta$ estimated from a full resolution image, from half res (b)
and from quarter res images (c). The derived $\alpha$ map is shown in (e) when
using full res $\beta$, (f) when using upsampled half res $\beta$ and (g) when
using upsampled quarter res $\beta$. The estimated $\alpha$ maps are consistent
through the scales (SSIM(e,f)=0.9768; MAD(e,f)=0.0085) and (SSIM(e, g)= 0.9544;
MAD(e,f)=0.0128).

Comparisons at full res, half res and quarter res for the 26 training images are
reported in Table~\ref{table:diffres} and confirm that estimating $\beta$ at lower
resolutions can be efficiently used for pulling an equivalent quality $\alpha$
map at full resolution.

\subsection{Running Time} The complexity of our approach is similar to 
closed-form matting. The main difference is that closed-form matting requires a
$5\times 5$ stencil for $\alpha$, whereas we only need a 5-stencil. On the other
hand, in closed-form matting we only manipulate a scalar ($\alpha$), whereas in
our approach we manipulate $\beta$ and $\x\x^t$, which are of dimensions
$4\times 1$ and $4 \times 4$. In this paper, the algorithms compared are
implemented in MATLAB using exact solvers and are not optimised for speed. We
believe however that our approach has the potential to offer better performances
as it requires less spatial access than closed-form, and is also better suited
for multi-grid iterative solvers because of the smoothness of $\beta$.

\subsection{Limitations}
Our method suffers from the same limitations as with closed-form matting. The
main issue is that the linear model of \eq{eq:linearmattingequation} is not
always a good approximation. This is especially true when the background is
cluttered. Also, as it is essentially a diffusion process, the geometry and
sparness of the trimap has a key influence on the results. It is typically
difficult to pick up isolated hair strands. Using sampling techniques as a
pre-process helps reducing the issue.

\begin{table}[t]
  \centering
    \begin{tabular}{|c|c|c|c|}
      \hline 
       SSIM mean & SSIM std & SAD mean & SAD std \\
      \hline
      \hline
       0.9751 & 0.02 & 0.0052 & 0.0051 \\
      \hline
    \end{tabular}
    \caption{ Difference between closed-form matting
       and our approach for the 26 training images
      of~\cite{alphamatting.com}.}
      \label{table:diff}
\end{table}

\begin{table}[t]
  \centering
    \begin{tabular}{|c|c|c|c|c|}
      \hline 
      & SSIM mean & SSIM std & SAD mean & SAD std \\
      \hline
      \hline     
      \sfrac{1}{2} res & 0.9441 & 0.0437 & 0.0118 & 0.0101 \\
      \hline
      \sfrac{1}{4} res & 0.9138 & 0.0596 & 0.019 & 0.0165\\
      \hline
    \end{tabular}
    \caption{Difference between our approach at full res and half res, full res
      and quarter res. The results are compiled using the 26 training images
      of~\cite{alphamatting.com}.  }
      \label{table:diffres}
\end{table}

\section{Conclusion}

We have proposed an alternative presentation for the closed-form matting
Laplacian. This alternative framework is built around a direct manipulation of
the linear matting model parameters $a$ and $b$. The resulting transparency maps
are very similar to the one obtained with closed-form matting. We have shown
however that our framework yields better expositions of the spatial and unary
priors and that the smoothness of the model parameters allow us to efficiently
scale up transparency maps to higher resolutions.

\bibliographystyle{IEEEbib} \bibliography{icip2016}

\end{document}